\documentclass{article}

\usepackage{amsmath}
\usepackage{PRIMEarxiv}
\usepackage{booktabs}
\usepackage[utf8]{inputenc} 
\usepackage[T1]{fontenc}    
\usepackage{hyperref}       
\usepackage{url}            
\usepackage{booktabs}       
\usepackage{amsfonts}       
\usepackage{nicefrac}       
\usepackage{microtype}      
\usepackage{lipsum}
\usepackage{fancyhdr}       
\usepackage{graphicx}       
\graphicspath{{media/}}     

\title{An Analysis of Embedding Layers and Similarity Scores using Siamese Neural Networks
}

\author{
  Yash Bingi, Yiqiao Yin \\
  Department of Statistics \\
  Columbia University \\
  New York City, New York\\
  \texttt{ybingi20@gmail.com, yy2502@columbia.edu} \\
}

\begin{document}
\maketitle

\begin{abstract}
Large Lanugage Models (LLMs) are gaining increasing popularity in a variety of use cases, from language understanding and writing to assistance in application development. One of  the most important aspects for optimal funcionality of LLMs is embedding layers. Word embeddings are distributed representations of words in a continuous vector space. In the context of LLMs, words or tokens from the input text are transformed into high-dimensional vectors using unique algorithms specific to the model. Our research examines the embedding algorithms from leading companies in the industry, such as OpenAI, Google's PaLM, and BERT. Using medical data, we have analyzed similarity scores of each embedding layer, observing differences in performance among each algorithm. To enhance each model and provide an additional encoding layer, we also implemented Siamese Neural Networks. After observing changes in performance with the addition of the model, we measured the carbon footage per epoch of training. The carbon footprint associated with large language models (LLMs) is a significant concern, and should be taken into consideration when selecting algorithms for a variety of use cases. Overall, our research compared the accuracy different, leading embedding algorithms and their carbon footage, allowing for a holistic review of each embedding algorithm.
\end{abstract}


\section{Introduction}
Since the Introduction of Large Language Models (LLMs) into computer science, they have been used to tackle many issues in a wide variety of academic fields.\cite{chang2023survey} They have been used in sentiment analysis problems, which involves determining the mood expressed in a piece of text, such as a tweet, review, or comment. LLMs, with their ability to understand context and nuances in language, can be trained to analyze and classify the sentiment of textual content, allowing for applications and products that can reply to users in a similar manner.\cite{SentimentAnalysis} Another important use case for LLMs is within the field of machine translation. LLMs have demonstrated remarkable capabilities in improving the accuracy and fluency of machine translation systems.\cite{karpinska2023large} Algorithms for language translation can become much more accurate and streamlined towards a wide variety of languages in a much more efficient manner. Lastly, LLMs have a huge potential in data generation and synthetic data creation.\cite{li2023synthetic} There are many things in the world that we would like to create machine learning and AI applications for, but we struggle with finding a large amount of data to test and run our models on. Large Language Models can help this field by creating synthetic data to match the small numbers of existing data we have. This can be done with all types of media, but most importantly, we can do this with textual data, which allows for LLMs to be most effective.

In order to carry out all of these use cases, there is a need for the LLMs to be able to understand and compute differences between sentences.\cite{bengio2000neural} In order to analyze sentiment, for example, they need to be able to know that words that are physically different have similar meanings, and they also need to know which sentences have similar meanings to each other. This same idea is true as well with translation devices, where it is imperative for differences in words to be picked up on. Lastly, and most importantly, with synthetic data, it is necessary to be able to analyze similarity between data points in order to follow the trend and continue to make fake data that meets the criteria of the application it is created for. In order to compute the named "differences" within the data, it is necessary to use embedding layers.\cite{jing2019survey} These algorithms are fundamental for converting input token sequences into continuous vector representations. This process involves the use of a learnable embedding matrix with rows representing the vectors for individual tokens. During training, the matrix is adjusted to capture meaningful semantic relationships between tokens. Each embedding algorithm used in our research is a little bit different, and has its own parameters.\cite{balikas2023comparative} Using these embedding algorithms, we can then calculate the similarity between words and sentences using similarity scores, which are complex equations that are again different by model, and can vary in their accuracy. Similarity scores in LLMs play a crucial role in tasks like information retrieval, document matching, and semantic understanding, contributing to the models' ability to capture and leverage contextual relationships within language.\cite{do2009robust} 

Currently, LLMs are quite good at generating responses to smaller, quicker, and more surface level questions. This was acceptable with the initial stages of LLMs, but with the growing demand for complex responses and deep answers for many of the applications above, LLMs have been a little more unreliable. In order to fix this issue, Retrieval-augmented Generation (RAG) has emerged as a leading tool to use in combination with LLMs.\cite{lewis2020retrieval} Retrieval-augmented generation is an approach in natural language processing that combines information retrieval and text generation. In this method, relevant information is first retrieved from a set of documents or a knowledge base based on the input context or query. Subsequently, a language model or generator uses this retrieved information to produce a coherent and accurate response based on the content and the sentiment of the question. This method is advantageous in tasks such as question answering and dialogue systems, where incorporating external knowledge is crucial for generating informed and contextually accurate outputs. Additionally, it performs very well when applied to many other use cases like the ones shown above. By seamlessly integrating retrieved content into the generation process, retrieval-augmented generation aims to strike a balance between structured information and creative language generation capabilities of the model, resulting in more contextually rich and unique responses. With the utilization of software like RAG, problems and questions asked by users will be answered with much more in detail and nuanced responses, which is not only better for the user, but also better in product development situations with things like language apps and synthetic data creators. \cite{cai2022recent}

In order to utilize RAG, it is necessary to be able to match the similarity between two sentences. This will allow RAG to more effectively search the knowledge base from which it is tasked to borrow information from. By more effectively being able to match sentence similarities, it will be easier for the algorithm to pull out relevant pieces of information and provide users with a wider variety of information that goes into a much deeper level of understanding than before. Our research focuses on this piece of similarity, analyzing the current leading similarity scores of a variety of companies and their effectiveness in correctly distinguishing the difference between sentences. With our research, we hope to find a model that can accurately distinguish whether or not 2 pieces of media, in our case sentences, are similar based off of their similarity scores. Using our results, similarity scores can be accurately used in the RAG process, which will allow for greater accuracy and depth in the functionality of LLMs.

\section{Methods and Materials}
The overarching goal of this research is to analyze the embedding algorithms and the accuracy of similarity scores among a variety of different leading algorithms. Through this research, we hope to provide information on the pros and cons of each embedding algorithm, allowing them to be able to use whichever algorithm they see fit for their application purposes. Along with analyzing the algorithms, our research shows the ecological effect on these models, ensuring that the use of any algorithms keeps the environmental impact in consideration. Overall, this paper aims to holistically analyze the embedding algorithms and similarity scores of OpenAI, PaLM, and BERT embeddings using a variety of different algorithms and carbon footage calculators.

\subsection{Data Set}
The data set used in our research was brought up after careful consideration over a variety of factors. The first factor that we had to take into consideration was to pick a data set with textual data. Without textual data, it would be impossible to use our embedding algorithms to tokenize and build vectors among each sentence. The next requirement of our data set was for it to have matching pairs. In order to test the accuracy of the different models we were planning on using, it was imperative for the data to have sentence pairs that were similar in meaning when looked at by a human, but different in physical wording to test the accuracy of our embedding algorithms and similarity scores. 

After scanning through many different databases for information online, we eventually landed on a set of commonly searched medical questions on Google. \cite{mccreery2020effective} This data set consists of 3048 similar and dissimilar medical question pairs hand-generated and labeled by Curai's doctors. Doctors with a list of 1524 patient-asked questions randomly sampled from the publicly available crawl of HealthTap. Each question results in one similar and one different pair through the following instructions provided to the labelers:

Rewrite the original question in a different way while maintaining the same intent.

Restructure the syntax as much as possible and change medical details that would not impact your response.

Come up with a related but dissimilar question for which the answer to the original question would be WRONG OR IRRELEVANT. Use similar key words.

The first instruction generated a positive question pair (similar) and the second generated a negative question pair (different).

Each piece of data created from this data set contained three different rows, or labels. The first row contained the first question, phrased in common English and a commonly searched question regarding public health and medicine on Google. The second row then contained another question. This question was phrased in the same level of English and was either similar in meaning to the first question or different from the first question. No matter what the case was, the second question was always worded differently in a way where it would not be optimal to just count the differences among wording in each sentence. Lastly, the third row would contain the label, which was marked as either a 1 or a 0. A 1 meant that the 2 sentences were similar in meaning, and a 0 meant the opposite, that the 2 sentences did not in fact mean the same thing. Table \ref{tab:data} below shows an example of the data that was used in this experiment. 

\begin{table}[h]
\centering
\begin{tabular}{@{}|l|l|l|@{}}
\toprule
\multicolumn{1}{|c|}{\textbf{Question\_1}}                                                                      & \multicolumn{1}{c|}{\textbf{Question\_2}}                                                                                                                            & \multicolumn{1}{c|}{\textbf{Label}} \\ \midrule
\begin{tabular}[c]{@{}l@{}}After how many hour from drinking \\ an antibiotic can I drink alcohol?\end{tabular} & \begin{tabular}[c]{@{}l@{}}I have a party tonight and I took my \\ last dose of Azithromycin this \\ morning. Can I have a few drinks?\end{tabular}                  & 1                                   \\ \midrule
\begin{tabular}[c]{@{}l@{}}After how many hour from drinking \\ an antibiotic can I drink alcohol?\end{tabular} & \begin{tabular}[c]{@{}l@{}}I vomited this morning and I am not \\ sure if it is the side effect of my \\ antibiotic or the alcohol I took last night...\end{tabular} & 0                                   \\ \midrule
\begin{tabular}[c]{@{}l@{}}Am I over weight (192.9) \\ for my age (39)?\end{tabular}                            & \begin{tabular}[c]{@{}l@{}}I am a 39 y/o male currently weighing \\ about 193 lbs. Do you think I am \\ overweight?\end{tabular}                                     & 1                                   \\ \midrule
\begin{tabular}[c]{@{}l@{}}Am I over weight (192.9) \\ for my age (39)?\end{tabular}                            & \begin{tabular}[c]{@{}l@{}}What diet is good for losing weight? \\ Keto or vegan?\end{tabular}                                                                       & 0                                   \\ \bottomrule
\end{tabular}
\caption{First four data points in data set}
\label{tab:data}
\end{table}

\subsection{Embedding Layer}
To start off our research with the data mentioned above, it was imperative to start by embedding our data into workable vectors. Embedding layers in neural networks, including Large Language Models (LLMs), serve a pivotal role in the conversion of discrete input, such as individual words or tokens, into continuous vector representations. At the core of this process is the embedding matrix, a learnable parameter within the layer. This matrix, initialized with random values, has rows corresponding to unique tokens in the model's vocabulary. During operation, the embedding layer performs a lookup operation for each input token, retrieving the associated row from the embedding matrix. These retrieved rows constitute the continuous vector representations, or embeddings, for the respective tokens. This is the general process of the embedding layers, and there are differences among each individual embedding algorithm that incorporate position and context of sentences. This allows different models to be more or less effective in specific situations. 

The first algorithm that was used in our research was the the BERT Embedding Algorithm. BERT's embedding algorithm distinguishes itself through its contextualized approach, employing bidirectional processing and a transformer architecture to capture nuanced language semantics. This gives it value in more complex English language settings, making it a valuable option for use cases that tend to use more advanced levels of English. Additionally, in contrast to traditional fixed word embeddings, BERT generates embeddings that are context-dependent, influenced by the entire context of a word within a sentence. This allows embeddings to change based on word meanings that could change based on the context of the sentence. The model is also pre-trained on a vast corpus using a masked language model objective, allowing it to predict masked words based on surrounding context and learn rich contextualized representations. BERT's large-scale pre-training on extensive data sets, including BooksCorpus and English Wikipedia, enhances its capacity to grasp diverse linguistic patterns and world knowledge. This makes it a great option for more diverse English patterns and information tasks from world settings. Furthermore, BERT embeddings can be fine-tuned for specific downstream tasks, making them adaptable to various natural language processing applications. Overall, the model's open-source availability and diversity in potential use cases have contributed to its widespread adoption and impact in the research and development community.

The next algorithm that was used in our research was the OpenAI Embedding Algorithm. This algorithm differs from traditional embeddings due to their unprecedented scale, massive size with 175 billion parameters, and extensive pre-training on diverse internet text. This allows the embedding algorithm to be much more effective and efficient on a wider variety of text. The contextual understanding is a standout feature, facilitated by a transformer architecture with attention mechanisms across multiple layers, enabling the model to grasp contextual relationships between words. This is good because not only does the model create vectors based on individual words, it incorporates the combination of words and the contextual meaning of words in its embeddings. Additionally, OpenAI's remarkable few-shot and zero-shot learning capabilities allow it to perform tasks with minimal examples or generalize to unseen tasks. This makes it a good choice for developers looking to embed fairly unique and generally complex pieces of data, like the ones we were working with. Overall, these factors collectively make OpenAI's embeddings distinctive, enabling versatile applications in a variety of natural language processing tasks.

The final algorithm that was used was Google's PaLM embedding algorithm. This algorithm distinguishes itself through its contextual understanding of sentences, capturing nuanced meaning and variations in context. Again, this allows for even the same words to have different embeddings based on the context of the sentence and what is going on. Unlike traditional word embeddings, PaLM generates fixed-size vectors for entire sentences, offering versatility across a spectrum of natural language processing tasks without the need for task-specific adaptations. By keeping the embeddings the same size, applications and flexibility with the possibilities of the algorithm greatly increases. Also, with support for multiple languages, PaLM extends its applicability to diverse linguistic contexts. This makes it the ideal model for text embeddings where different languages other than English are incorporated. Lastly, pre-trained on a large and diverse data set, PaLM's ongoing updates and commitment to incorporating advancements in NLP research underscore its position as a versatile and continually evolving tool in the field.

\subsection{Similarity Scores}
The next step in our research was to use the different embeddings to calculate the similarity scores between each data point. Using iteration, we looped through a program and individually calculated the similarity score of each embedding layer using its respective similarity score. In addition, we calculated using cosine similarity as a benchmark. The purpose of this part of the research process was to measure the performance of each of the embedding layers in order to properly identify which embedding algorithm was most effective for our training situation and in general with all types of contextual data.

The first type of score we used to calculate the similarity scores were cosine similarity. This was a very early form of measuring similarity between two pieces of text, and was used in our experiment as a baseline measure. In the realm of Large Language Models (LLMs), cosine similarity serves as a fundamental metric for evaluating the similarity between vector representations of words, sentences, or documents. These vector representations, often derived from the embeddings generated by LLMs, map textual elements into high-dimensional spaces where each dimension represents a semantic feature. Cosine similarity calculates the cosine of the angle between two vectors, with a resulting score ranging from -1 (indicating dissimilarity) to 1 (indicating similarity). A score of 0 implies orthogonality, meaning there is no similarity. This metric is particularly useful in tasks like text similarity assessment, where understanding the semantic relatedness of textual elements is crucial. This was what our experiment sought out to do, which is why we used it as a baseline quantity. By quantifying the cosine of the angle between vectors, cosine similarity provides a measure that considers both the magnitude and direction of the embeddings, offering valuable insights into the semantic relationships within the language model's learned representations. Formula \ref{eqn:cosine} that was used to calculate the cosine similarity between each vector is relatively straightforward, and is shown below.

\begin{equation}
\label{eqn:cosine}
\text{cosine\_similarity} = \frac{\mathbf{A} \cdot \mathbf{B}}{\|\mathbf{A}\| \cdot \|\mathbf{B}\|}
\end{equation}

In addition to using cosine similarity, the algorithm for BERT similarity was used next. This algorithm was most compatible with the BERT embeddings that we already have, which is why it was preferred to be used for those specific embeddings. The algorithm used is similar to cosine embedding, but it is specifically tailored for the BERT embeddings that were created in the previous section. Equation \ref{eqn:BERT} shows BERT's similarity score algorithm.

\begin{equation}
\label{eqn:BERT}
 \text{similarity\_score} = \frac{\text{BERT}(t_1) \cdot \text{BERT}(t_2)}{\|\text{BERT}(t_1)\| \cdot \|\text{BERT}(t_2)\|}   
\end{equation}

Here, the dot symbol represents the dot products of the BERT embeddings and the 2 values in the numerator denote the Euclidean norms (lengths) of the BERT embeddings. As mentioned earlier, this formula is a standard cosine similarity calculation, where the dot product of the embeddings is normalized by the product of their magnitudes. The resulting score is a measure of the cosine of the angle between the vectors, providing a measure of similarity between the two text inputs. While the base formula is the same, fine-tuning for specific applications might involve additional considerations or modifications to this basic formula.

The next similarity scores utilized were OpenAI's similarity scores based on their embeddings. Similar to BERT, these similarity scores were most compatible with the embeddings from OpenAI, which is what we measured them with. Formula \ref{eqn:OpenAI} shows the embedding algorithm, which is the same as the first two but with different inputs being used. 

\begin{equation}
\label{eqn:OpenAI}
    \text{similarity\_score} = \frac{\text{OpenAI}(t_1) \cdot \text{OpenAI}(t_2)}{\|\text{OpenAI}(t_1)\| \cdot \|\text{OpenAI}(t_2)\|}
\end{equation}

This has the same values and calculations as the previous algorithms, with the dot representing the dot product between the two vectors and the magnitudes of each vector representing the length.

After OpenAI, we then transferred over to use the PaLM embeddings. This was again only compatible with the Google PaLM embeddings, so that is what we tested it on for maximum efficiency. The algorithm was the same for this model as well, just with the changed vectors for embedding. Equation \ref{eqn:PaLM} shows the Google PaLM similarity score equtation. 

\begin{equation}
\label{eqn:PaLM}
    \text{similarity\_score} = \frac{\text{PaLM}(t_1) \cdot \text{PaLM}(t_2)}{\|\text{PaLM}(t_1)\| \cdot \|\text{PaLM}(t_2)\|}
\end{equation}

And as the same as the last 3 equations, the numerator calculated the dot product between the 2 equations on top while the denominator calculated the magnitude and length of each of the vectors.

Overall, each model was uniquely trained for it's specific embedding algorithm and was used uniquely based on the possible inputs that could be used. 

\subsection{Siamese Neural Network}
In order to try and enhance the power of the similarity scores, we looked to build a Siamese Neural Network using the embeddings. The main purpose of this neural network in our research was to try and boost the performance of our similarity scores and see which algorithms were resistant to this change. Based on the results with the Siamese Neural Network, we will be able to easily identify which models increased in accuracy, which stayed the same, and which models were hindered in progress based on what the results show. 

A Siamese Neural Network is a unique architecture used primarily for learning similarity or relationship between two comparable inputs. In our research problem and context, it was employed to determine the similarity between the tokenized sentences in each row of our data set. The core concept of a Siamese Neural Network is to map input features into an abstract space where similar items are modeled to be closer to each other than dissimilar items.

For a Siamese Neural Network dealing with sentences, the process begins with the tokenization of sentences. Our research used 3 different types of these tokenizing algorithms. The first one we used was BERT, the next one was OpenAI, and the last one was PaLM. Everything else about the models were kept the same in order to only measure the differences in performance based off of each of the individual embeddings.

The embedded vectors were then fed into twin networks, which are two identical neural networks with the same architecture and shared weights. This symmetry ensures that the networks process the two inputs in the same way, maintaining consistency in feature extraction. The outputs of these twin networks are feature vectors which represent the sentences in a transformed feature space. The next step was to measure the similarity between these feature vectors through a specific algorithm picked out to be used in the Siamese Neural Network. Overall, the general algorithm for the Siamese Neural Network is shown in equation \ref{eqn:SNN}.

\begin{equation}
\label{eqn:SNN}
    \hat{y} = \sigma(\text{FinalLayer}(\text{Concatenate}(\text{Embed}(s_1), \text{Embed}(s_2))))
\end{equation}

Here, Concatenate represents the operation that concatenates the embeddings of the two sentences, and FinalLayer denotes the final dense layer that produces the similarity score.

In training a Siamese Neural Network, a crucial aspect is the loss function. For this model, Binary Cross-Entropy (BCE) is employed as the loss function. Binary cross-entropy loss is often preferred in binary classification tasks, such as those involving Siamese neural networks for sentence similarity, due to its alignment with the nature of the problem. This loss function is well-suited for scenarios where the goal is to predict whether an instance belongs to one of two classes, offering sensitivity to predicted probabilities and a probabilistic interpretation. Our data contained only two classes, which is why we decided to use it for this project. Additionally, its emphasis on penalizing confident misclassifications promotes the development of well-calibrated models. Moreover, the simplicity and computational efficiency of binary cross-entropy make it a popular and accessible choice. The BCE function is shown below in equation \ref{eqn:loss}. 

\begin{equation}
\label{eqn:loss}
\text{Binary Cross-Entropy Loss} = - \frac{1}{N} \sum_{i=1}^{N} \left( y_i \cdot \log(\hat{y}_i) + (1 - y_i) \cdot \log(1 - \hat{y}_i) \right)
\end{equation}

By minimizing the Binary Cross-Entropy loss during training, the Siamese Network learns to output similarity scores that closely align with the true similarity of the sentence pairs. This makes the Siamese Neural Network a powerful tool for tasks like semantic similarity analysis, which is exactly what we are using it for in our experiment.

We created a Siamese Neural Network and fit it to approximately 75\% of our data. Then, we used the remaining data as test data and obtained our results. Figure \ref{fig: SNN_Chart} below shows the model and the path that was created to fit our data to the specific Siamese Neural Network we created.

\begin{figure}[h]
\centering
\includegraphics[scale=0.75]{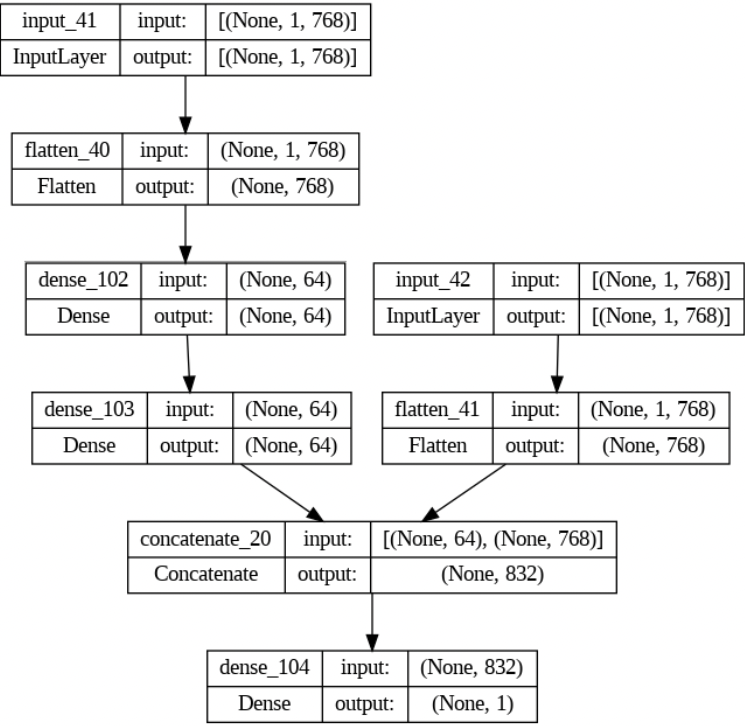}
\caption{Model of the Siamese Neural Network used}
\label{fig: SNN_Chart}
\end{figure}

\section{Results}
Overall, after calculating the similarity scores of each of the different algorithms and embedding layers and then averaging out the total success rate, we were able to compare the average success rate among each of the different similarity score measures.

Figure \ref{fig: Similarity_Chart} below shows a box plot of the similarity scores between the different similarity score algorithms. The first box shows the accuracy of the cosine similarity, the next box shows the accuracy of the BERT Similarity scores, and so on. The OpenAI and the PaLM similarity scores are also displayed on the box chart between the different embedding algorithms. 

\begin{figure}[h]
\centering
\includegraphics[scale=0.5]{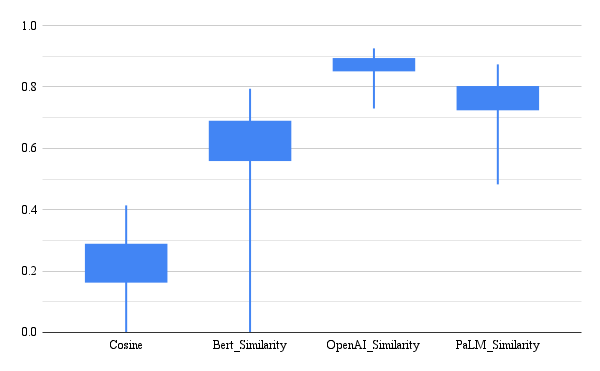}
\caption{Box Chart showing average accuracy of similarity scores}
\label{fig: Similarity_Chart}
\end{figure}

As we can see in the graph, cosine similarity was quite low in all metrics. This is because the algorithm used to embed and then compute the similarity scores is quite basic. It takes any two words that are present in both sentences and counts that towards similarity. Even though two words may be in different positions within the sentence or many have different meanings in the context of the sentence, these nuances are not accounted for and are therefore seen as similar. As a result, the average predicted probability was a little over 0.2, and the standard deviation range generally stayed within the 0-0.4 range. 

The next embedding algorithm used was BERT, and the BERT similarity scores were calculated based on the findings from this table. As we can see, the minimum accuracy for Bert is extremely low, almost as low as the values we were getting with our cosine similarity. However, other than those few outliers, the performance of BERT was generally quite good and it had a median accuracy of around 0.6. While this isn't ideal in any way, it is still considerably better than the cosine similarity. This could be attributed to the fact that it's embedding system is able to pick up on complex language structure and it is able to account for complex nuances within the English language. The standard deviation was also quite high, however, and the upper quartile had a pretty big range. Overall, BERT shows great improvement from the previous cosine similarity but still struggled with certain sentences, making it adequate but not efficient in the sentence similarity process. 

OpenAI was the next bar in the graph, and it performed exceedingly well given our data set and the task at hand. The median value of the box for OpenAI is very high, and it sits just under 0.9. In addition to the extremely high median accuracy, the standard deviation is also much lower than any of the other similarity algorithms. The upper quartile especially is extremely small and with the exception of a few outliers bringing the lower quartile down, the accuracy is well clustered at a high value. The reason for the high performance of OpenAI's model could be attributed to a variety of factors, such as its complex understanding of the human language and its ability to navigate the positioning of words within sentences. It assigns different vectors to the words based on their position within the text, which makes it much more accurate and able to understand the meaning of sentences much more effectively. 

Lastly, PaLM similarity scores also showed quite a high value of accuracy when paired with the right embedding algorithm. Its average accuracy was again much higher than the cosine values, and was also slightly higher than the accuracy of the BERT similarity scores. The standard deviation was higher than OpenAI's, but it was generally smaller than the other two measures shown. The lower quartile had quite a high spread while the upper quartile didn't have as much. This is a trend among all the different similarity scores, and is likely due to outliers in the accuracy of certain sentences that may have been harder to tokenize. 

The next step in our research was to then measure the results of our Siamese Neural Networks. We ran 3 neural networks, each one using a different algorithm. Each of the neural networks were base models and were kept exactly the same as each other to ensure that there would be no confounding variables in the process. Figure \ref{fig: SNN_Chart} shows the box plots of the accuracy of the Siamese Neural Network models when inputted with the different similarity scores. Each model was training with a 75-25 train test split, which is the general industry standard when dealing with these types of data.

\begin{figure}[h]
\centering
\includegraphics[scale=0.5]{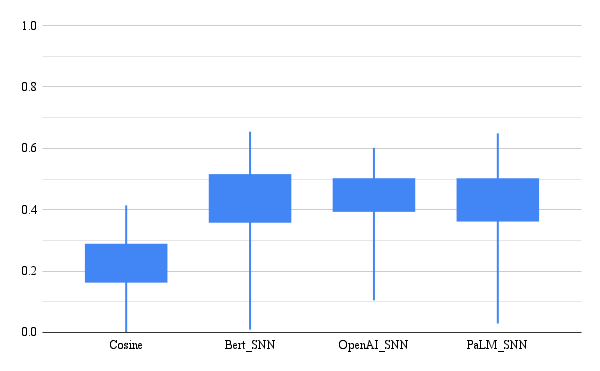}
\caption{Box Chart showing average accuracy of the base Siamese Neural Network}
\label{fig: SNN_Chart}
\end{figure}

As you can see in this chart, we kept the cosine similarity scores in as a benchmark to compare the other base models to. The reason that the accuracy is low for each of the other boxes is because we used a quite basic Siamese Neural Network. The main purpose of this step in the experiment was not to aim for an extremely high accuracy through rigorous tuning and optimization of the model. Instead, we were looking to compare the different embedding algorithms and their general compatibility with neural network models in general. We were looking for the most adaptable embedding algorithms, and this method of testing with base models allows us to see which embedding algorithm works the best when manipulated and fit to neural network models like the one we have created.

As we can see in the graph shown above, there is not much seperating each of the algorithms and their Siamese Neural Network adaptations. While BERT does have the highest maximum of the four algorithms, it also has the lowest accuracy among the four. OpenAI again has a low standard deviation, which makes it good generally reliable when looking to apply embedding to a certain problem. Lastly, PaLM has a high range similar to BERT and OpenAI, but the standard deviation seems to be in the middle of OpenAI and BERT, making it the middle algorithm between the two.

\subsection{Carbon Footage}
The last step of our research project was to compute the carbon footage of each of the embedding algorithms and similarity scores. The carbon footprint of AI models is a critical consideration as the field continues to evolve.\cite{lottick2019energy} The energy-intensive nature of training large-scale models, such as deep neural networks like the ones being used in our research and in many other tasks around the world, contributes significantly to carbon emissions. The computational demands of training these models, which are often performed in data centers powered by non-renewable energy sources in large-scale systems, result in substantial environmental impact. Furthermore, the trend toward developing ever-larger models amplifies these concerns, necessitating a careful examination of the sustainability of AI practices. Beyond the training phase, the operational energy use of deployed AI models adds to their long-term environmental footprint.

In order to take these concerns into account, our research aims to check the carbon footage of each of our different models. We want to ensure that the models and similarity scores outputted are environmentally friendly and have the potential to be used in the long run. In order to measure this, we measured the carbon footage per epoch, or training run, of the model. Additionally, we then measured the total carbon footage emitted by each of the models. Figure \ref{tab:Carbon} shows the results of our experiment.

\begin{table}[h]
\centering
\begin{tabular}{@{}|l|c|c|c|@{}}
\toprule
\multicolumn{1}{|c|}{\textbf{}} & \multicolumn{1}{l|}{\textbf{BERT\_SNN}} & \multicolumn{1}{l|}{\textbf{OpenAI\_SNN}} & \textbf{PaLM\_SNN}         \\ \midrule
\textbf{Footage Per Epoch}      & 4.4*10\textasciicircum{}-5              & 5.3*10\textasciicircum{}-5                & 5.0*10\textasciicircum{}-5 \\ \midrule
\textbf{Footage Per Model}      & 8.4*10\textasciicircum{}-5              & 1.2*10\textasciicircum{}-3                & 7.9*10\textasciicircum{}-5 \\ \bottomrule
\end{tabular}
\caption{Carbon footage per epoch and total for each SNN model}
\label{tab:Carbon}
\end{table}

As we can see in the experiment, BERT has the lowest footage per epoch of carbon emissions. This makes sense because its embedding algorithm is generally the least robust and the most efficient out of the three. While this does impact the performance of the model slightly, it is useful in keeping the carbon emissions low. The next lowest carbon emissions per epoch was PaLM embeddings. This is consistent with the performance of the models, and due to its slightly more complex algorithm and computing power necessary, it ranks as the second highest producer of carbon. Lastly, OpenAI produces the most carbon emissions per epoch. This is likely due to its extremely robust and complex algorithm. The computing power necessary for the algorithm is much higher than the other two and more energy is used, causing a higher rate of carbon emissions. 

When it comes to the total emissions, however, the trend changes a little bit. PaLM ends up having the lowest carbon emissions, meaning that although it has very high levels of carbon, it's carbon emissions level off as we go through the epochs. This makes it really good for scenarios when we are training with a high number of epochs, because the additional carbon footprint for each epoch levels off and eventually doesn't significantly contribute to the total carbon emissions. The next algorithm is Bert, which almost doubles exactly in carbon emissions. This is not good because it makes it unscalable to larger applications and cannot be used with a lot of epochs in the training algorithm. Given its relatively low performance from the other two as well, you wouldn't be getting too much out of this algorithm for the high level of carbon emissions it outputs. Lastly, the highest carbon emissions come from OpenAI. The total emissions are extremely high and much greater than any of the other two models. While it is extremely robust and high performing in a variety of applications, the extreme amount of carbon emissions released make it less than ideal for many applications. In many cases, it might be sensible to take a slight drop in accuracy with the PaLM or BERT models in exchange for a much more environmentally friendly model that doesn't cause as much pollution.

\section{Conclusion}
In conclusion, each of the models and similarity scores above have their own pros and cons, and should be used based on the situation that you would need. First off, the BERT embedding algorithm showed an increase in similarity scores from the baseline cosine similarity scores. The reason for this increase could be attributed to BERT's ability to generate contextualized word embeddings and capture intricate linguistic nuances by considering both left and right context within a sentence. Leveraging a bidirectional transformer architecture, BERT excels in understanding complex sentence structures and semantic relationships, contributing to its overall success in the task we tested it on. Additionally, the model is pre-trained on extensive and diverse datasets, allowing it to learn rich and generalizable language representations. Although it didnt' have the highest performance in the end, the carbon footage was quite low, making it a great option for quicker, simpler text similarity tasks where a less robust but still efficient model is necessary. 

The next embedding algorithm we looked at was OpenAI's model. OpenAI showed extremely high performance when analyzed with the problem given, and did very good on the task given. Additionally, it had a very low standard deviation, making it reliable on a wide variety of problems. The reasons for OpenAI's overall immense success could be attributed to its generative prowess and contextual understanding. OpenAI's ability to generate coherent and contextually relevant text contributes to the high quality of the embeddings it produced in our research. Pre-trained on a vast and diverse data set, OpenAI captures intricate patterns and relationships within language, resulting in embedding algorithms that generalize effectively across the variety of different sentences we used. Overall, OpenAI proved to be the highest performing algorithm of the three due to its robust algorithm for embedding and its ability to understand relationships between words in the sentences and how the meaning of the sentence was affected by that. The biggest limitation of OpenAI, however, was the fact that it used up so much carbon and released the most emissions per epoch and in total. Due to it's robust algorithm, it was not scalable at all and ended up producing multiples of the other algorithms in terms of carbon emissions.

Lastly, we looked at Google's PaLM embedding algorithm. This algorithm was again in the middle of the pack when it comes to performance, and lagged just behind OpenAI in terms of overall accuracy. However, it is still a great algorithm to use because of its efficiency in capturing semantic relationships between words. The continuous vector representations generated by the algorithms, often available through pre-trained models, excel in measuring word similarity and finding meaningful associations, which is what we tested it on through our research. In addition to their good performance and quick run times, they are also extremely scalable. Of the three models, they had the lowest total carbon emissions. Even though the first epoch had a large carbon footprint, this was nullified through the extremely low footprint added for additional epochs being run. The total carbon footprint was the lowest out of the three, making this algorithm the best one by far for a great blend of accuracy and eco-friendliness. The scalable nature of this model makes it ideal for larger scale problems where a greater amount of data needs to be processed and embedded.

\section{Further Research}
In the future, we would be interested in comparing different embedding algorithms that are newer and may be just starting to pick up. This could be beneficial because it would allow us to determine which algorithms have the potential to be used in larger scale applications and which algorithms may have potential to become mainstream in the future. Additionally, by comparing the accuracy and the carbon emissions of each of the different newer models with the already existing models, we will be able to understand the pros and cons of the already existing models and what each one excels in based on how it compares with the other models.

Another point of interest in the future would be to maximize the efficiency of the Siamese Neural Network that was used in the research. We could spend time optimizing the parameters and fine tuning the models in order to achieve the highest possible performance. By doing this, we might be able to create an extremely high performance similarity score algorithm that is better than the individual algorithms created by each of the companies. This would be ideal because it would help create a highly accurate embedding and similarity algorithm that could be used with smaller and more individualized sets of data.

\bibliographystyle{unsrt}  
\bibliography{references}

\end{document}